\documentclass[10pt,twocolumn,letterpaper]{article}

\usepackage{cvpr}
\usepackage{graphicx}
\usepackage{verbatim}
\usepackage{footmisc}
\usepackage{times}
\usepackage{epsfig}
\usepackage{graphicx}
\usepackage{amsmath}
\usepackage{amssymb}
\usepackage{algorithm2e}
\usepackage[dvipsnames]{xcolor}
\usepackage{color}
\usepackage{enumitem}
\setlist{nolistsep}
\newsavebox\dotbox
\sbox{\dotbox}{\(\displaystyle\bigodot\)}
\DeclareMathOperator*{\bigcdot}{\raisebox{0pt}[\ht\dotbox][\dp\dotbox]{\(\boldsymbol{\cdot}\)}}
\DeclareMathSizes{10}{9}{6}{6}

\SetAlFnt{\small\sffamily}
\usepackage[pagebackref=true,breaklinks=true,letterpaper=true,colorlinks=true,bookmarks=false,backref]{hyperref}

 \cvprfinalcopy 

\def\cvprPaperID{2451} 

\ifcvprfinal\fi
\begin{document}
\pagenumbering{gobble}
\title{Switching Convolutional Neural Network for Crowd Counting}

\author{Deepak Babu Sam\thanks{Equal contribution}  \qquad Shiv Surya\footnotemark[1] \qquad R. Venkatesh Babu \\ Indian Institute of Science\\
 Bangalore, INDIA 560012 \\ {\tt\small bsdeepak@grads.cds.iisc.ac.in, shiv.surya314@gmail.com, venky@cds.iisc.ac.in}}

\maketitle
\begin{abstract}

We propose a novel crowd counting model that maps a given crowd scene to its density. Crowd analysis is compounded by myriad of factors like inter-occlusion between people due to extreme crowding, high similarity of appearance between people and background elements, and large variability of camera view-points. Current state-of-the art approaches tackle these factors by using multi-scale CNN architectures, recurrent networks and late fusion of features from multi-column CNN with different receptive fields. We propose switching convolutional neural network that leverages variation of crowd density within an image to improve the accuracy and localization of the predicted crowd count. Patches from a grid within a crowd scene are relayed to independent CNN regressors based on crowd count prediction quality of the CNN established during training. The independent CNN regressors are designed to have different receptive fields and a switch classifier is trained to relay the crowd scene patch to the best CNN regressor. We perform extensive experiments on all major crowd counting datasets and evidence better performance compared to current state-of-the-art methods. We  provide interpretable representations of the multichotomy of space of crowd scene patches inferred from the switch. It is observed that the switch relays an image patch to a particular CNN column based on density of crowd.

\end{abstract}


\section{Introduction}
Crowd analysis has important geo-political and civic applications. Massive crowd gatherings are commonplace at candle-light vigils, democratic protests, religious gatherings and presidential rallies. Civic agencies and planners rely on crowd estimates to regulate access points and plan disaster contingency for such events. Critical to such analysis is crowd count and density. 

\begin{figure}[!t]
 \includegraphics[width =0.48\textwidth, height=5.0cm]{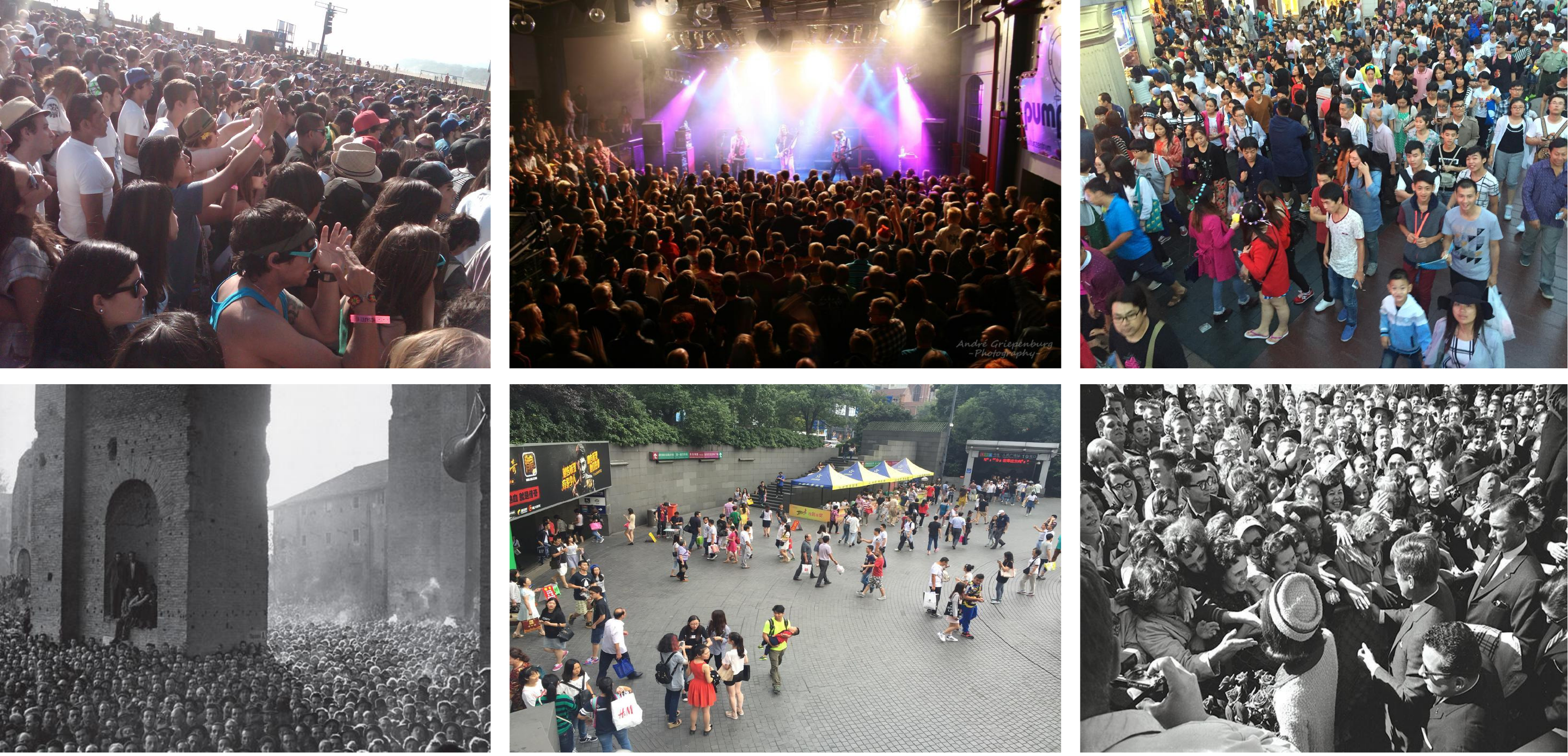}\hspace{0.3cm}%
 \caption{Sample crowd scenes from the ShanghaiTech dataset~\cite{zhang2016single} is shown.}

\label{mosaic_cc}
\end{figure}

In principle, the key idea behind crowd counting is self-evident: density times area. However, crowds are not regular across the scene. They cluster in certain regions and are spread out in others. Typical static crowd scenes from the ShanghaiTech Dataset~\cite{zhang2016single} are shown in  Figure~\ref{mosaic_cc}. We see extreme crowding, high visual resemblance between people and background elements (e.g. Urban facade) in these crowd scenes that factors in further complexity. Different camera view-points in various scenes create perspective effects resulting in large variability of scales of people.

Crowd counting as a computer vision problem  has seen drastic changes in the approaches, from early HOG based head detections~\cite{idrees2013multi} to CNN regressors~\cite{zhang2015cross,zhang2016single,onoro2016towards} predicting the crowd density. CNN based regressors have largely outperformed traditional crowd counting approaches based on weak representations from local features. We build on the performance of CNN based architectures for crowd counting and propose Switching Convolutional Neural Network (\textbf{Switch-CNN}) to map a given crowd scene to its density. 

Switch-CNN leverages the variation of crowd density within an image to improve the quality and localization of the predicted crowd count. Independent CNN crowd density regressors are trained on patches sampled from a grid in a given crowd scene. The independent CNN regressors are chosen such that they have different receptive fields and field of view. This ensures that the features learned by each CNN regressor are adapted to a particular scale. This renders Switch-CNN robust to large scale and perspective variations of people observed in a typical crowd scene. A particular CNN regressor is trained on a crowd scene patch if the performance of the regressor on the patch is the best. A switch classifier is trained alternately with the training of multiple CNN regressors to correctly relay a patch to a particular regressor. The joint training of the switch and regressors helps augment the ability of the switch to learn the complex multichotomy of space of crowd scenes learnt in the differential training stage. 

To summarize, in this paper we present:
\begin{itemize}
\item A novel generic CNN architecture, Switch-CNN trained end-to-end to predict crowd density for a crowd scene.
\item Switch-CNN maps crowd patches from a crowd scene to independent CNN regressors to minimize count error and improve density localization exploiting the density variation within a scene. 
\item We evidence state-of-the-art performance on all major crowd counting datasets including ShanghaiTech dataset~\cite{zhang2016single}, UCF\_CC\_50 dataset~\cite{idrees2013multi} and WorldExpo'10 dataset~\cite{zhang2015cross}.
\end{itemize}

\section{Related Work}
Crowd counting has been tackled in computer vision by a myriad of techniques. Crowd counting via head detections has been tackled by \cite{wu2005detection,wang2011automatic,viola2005detecting} using motion cues and appearance features to train detectors. Recurrent network framework has been used for head detections in crowd scenes by \cite{stewart2015end}. They use the deep features from Googlenet~\cite{szegedy2015going} in an LSTM framework to regress bounding boxes for heads in a crowd scene. However, crowd counting using head detections has limitations as it fails in dense crowds, which are characterized by high inter-occlusion between people.

In crowd counting from videos, \cite{brostow2006unsupervised} use image features like Tomasi-Kanade features into a motion clustering framework. Video is processed by \cite{rabaud2006counting}  into a set of trajectories using a KLT tracker. To prevent fragmentation of trajectories, they condition the signal temporally and spatially. Such tracking methods are unlikely to work for single image crowd counting due to lack of temporal information.

Early works in still image crowd counting like \cite{idrees2013multi} employ a combination of handcrafted features, namely HOG based detections, interest points based counting and Fourier analysis. These weak representations based on local features are outperformed by modern deep representations.
In \cite{zhang2015cross}, CNNs are trained to regress the crowd density map. They retrieve images from the training data similar to a test image using density and perspective information as the similarity metric. The retrieved images are used to fine-tune the trained network for a specific target test scene and the density map is predicted. However, the model's applicability is limited by fine-tuning required for each test scene and  perspective maps for train and test sequences which are not readily available. An Alexnet~\cite{krizhevsky2012imagenet} style CNN model is trained by \cite{wang2015deep} to regress the crowd count. However, the application of such a model is limited for crowd analysis as it does not predict the distribution of the crowd.   
In \cite{onoro2016towards}, a multi-scale CNN architecture is used to tackle the large scale variations in crowd scenes. They use a custom CNN network, trained separately for each scale. Fully-connected layers are used to fuse the maps from each of the CNN trained at a particular scale, and regress the density map. However, the counting performance of this model is sensitive to the number of levels in the image pyramid as indicated by performance across datasets.

Multi-column CNN used by \cite{boominathan2016crowdnet,zhang2016single} perform late fusion of features from different CNN columns to regress the density map for a crowd scene. In \cite{zhang2016single}, shallow CNN columns with varied receptive fields are used to capture the large variation in scale and perspective in crowd scenes. Transfer learning is employed by \cite{boominathan2016crowdnet} using a VGG network employing dilated layers complemented by a shallow network with different receptive field and field of view. Both the model fuse the feature maps from the CNN columns by weighted averaging via a 1$\times$1 convolutional layer to predict the density map of the crowd. However, the weighted averaging technique is global in nature and does not take in to account the intra-scene density variation. We build on the performance of multi-column CNN and incorporate a patch based switching architecture in our proposed architecture, Switch-CNN to exploit local crowd density  variation within a scene (see Sec \ref{scnn} for more details of architecture). 

While switching architectures have not been used for counting, expert classifiers have been used by \cite{sarvadevabhatla2016swiden} to improve single object image classification across depiction styles using a deep switching mechanism based on depiction style. However unlike \cite{sarvadevabhatla2016swiden}, we do not have labels (For eg: Depiction styles like "art" and "photo") to train the switch classifier. To overcome this challenge, we propose a training regime that exploits CNN regressor's architectural differences (See Section \ref{scnn})


\begin{figure}[!t]
 \includegraphics[width =0.48\textwidth, height=8.0cm]{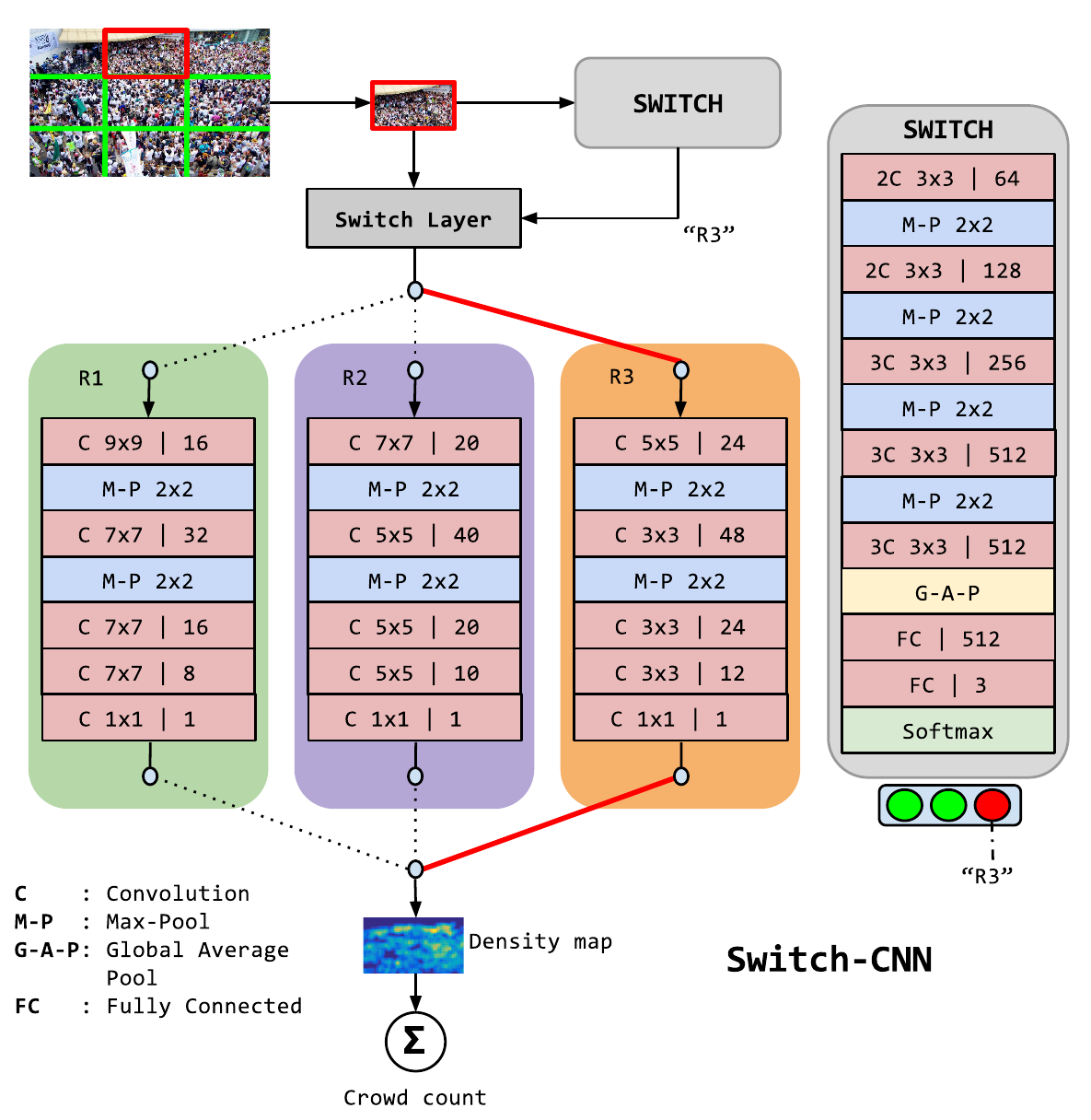}\hspace{0.3cm}%
 \caption{Architecture of the proposed model, \textbf{Switch-CNN} is shown. A patch from the crowd scene is highlighted in \textcolor{red}{red}. This patch is relayed to one of the three CNN regressor networks based on the CNN label inferred from \textbf{Switch}. The highlighted patch is relayed to regressor $R_3$ which predicts the corresponding crowd density map. The element-wise sum over the entire density map gives the crowd count of the crowd scene patch.}
\label{arch}
\end{figure}

\section{Our Approach}
Convolutional architectures like \cite{zhang2015cross,zhang2016single,onoro2016towards} have learnt effective image representations, which they leverage to perform crowd counting and density prediction in a regression framework. Traditional convolutional architectures have been modified to model the extreme variations in scale induced in dense crowds by using multi-column CNN architectures with feature fusion techniques to regress crowd density.

In this paper, we consider switching CNN architecture (Switch-CNN) that relays patches from a grid within a crowd
scene to independent CNN regressors based on a switch classifier.
The independent CNN regressors are chosen with different receptive
fields and field-of-view as in multi-column CNN networks to augment
the ability to model large scale variations. A particular CNN regressor
is trained on a crowd scene patch if the performance of the regressor
on the patch is the best. A switch classifier is trained
alternately with the training of multiple CNN regressors to correctly
relay a patch to a particular regressor. The salient properties that
make this model excellent for crowd analysis are (1) the ability
to model large scale variations (2) the facility to leverage local
variations in density within a crowd scene. The ability to leverage
local variations in density is important as the weighted averaging
technique used in multi-column networks to fuse the features is global
in nature.

\subsection{Switch-CNN}
\label{scnn}
\begin{algorithm}[!t]

\SetAlgoLined

\SetKwData{Left}{left}\SetKwData{This}{this}\SetKwData{Up}{up}
\SetKwFunction{Union}{Union}\SetKwFunction{FindCompress}{FindCompress}
\SetKwInOut{Input}{input}\SetKwInOut{Output}{output}
\Input{$N$ training image patches $\{X_{i}\}_{i=1}^{N}$ with ground truth density maps $\{D_{X_{i}}^{GT}\}_{i=1}^{N}$\\
 }
\Output{Trained parameters $\{\Theta_{k}\}_{k=1}^{3}$ for  $R_k$ and  $\Theta_{sw}$ for the switch}

\BlankLine

\BlankLine
Initialize $\Theta_{k}$ $\forall$ $k$ with random Gaussian weights\\
\BlankLine
Pretrain  $\{R_k\}_{k=1}^{3}$  for $T_{p}$  epochs\ : $R_k \leftarrow f_{k}(\bigcdot ; \Theta_{k})$ ;
\BlankLine
\color{BlueViolet}

\BlankLine
/*Differential Training for $T_{d}$ epochs*/\\
/*$C_{i}^{k}$ is count predicted by $R_k$ for input $X_i$*/\\
/*$C_{i}^{GT}$ is ground truth count for input $X_i$*/

\For {$t$ = 1 to $T_{d}$}
{
\For {$i$ = 1 to $N$}
{ 
$l_{i}^{best}=\underset{k}{\textrm{argmin}}|C_{i}^{k}-C_{i}^{GT}|$\;
 Backpropagate  $R_{l_{i}^{best}}$ and update $\Theta_{l_{i}^{best}}$\;
}
}

\color{OliveGreen}
\BlankLine
/*Coupled Training for $T_{c}$ epochs*/\\
Initialize $\Theta_{sw}$ with VGG-16 weights \;
\For {$t$ = 1 to $T_{c}$}
{
\BlankLine
\color{red}

/*generate labels for training switch*/\\
\For {$i$ = 1 to $N$}
{

$l_{i}^{best}=\underset{k}{\textrm{argmin}}|C_{i}^{k}-C_{i}^{GT}|$\;
}
$S_{train}=\{(X_i,l_{i}^{best}) \mid i \in[1,N]\}$\\
/*Training switch for 1 epoch*/\\
Train switch with $S_{train}$ and update $\Theta_{sw}$\;
\BlankLine
\color{BlueViolet}
/*Switched Differential Training*/  \\
\For {$i$ = 1 to $N$}
{
/*Infer choice of $R_k$ from switch*/ \\
$l_{i}^{sw}=\textrm{argmax }f_{switch}(X_{i};\Theta_{sw})$\;
Backpropagate  $R_{l_{i}^{switch}}$ and update $\Theta_{l_{i}^{sw}}$\;
\color{OliveGreen}
}

}

\color{black}
\caption{Switch-CNN training algorithm is shown. The training algorithm is divided into stages coded by color. \textbf{Color code index}: \textcolor{BlueViolet}{Differential Training}, \textcolor{OliveGreen}{Coupled Training}, \textcolor{red}{Switch Training}}
\label{algo}
\end{algorithm}

Our proposed architecture, Switch-CNN consists of three CNN regressors with different architectures and a classifier (switch) to select the optimal regressor for an input crowd scene patch. Figure~\ref{arch} shows the overall architecture of Switch-CNN. The input image is divided into 9 non-overlapping patches such that each patch is $\frac{1}{3}$\textsuperscript{rd} of the image.
For such a division of the image, crowd characteristics like density, appearance etc. can be assumed
to be consistent in a given patch for a crowd scene. Feeding patches as input
to the network helps in regressing different regions of the image
independently by a CNN regressor most suited to patch attributes like density, background, scale and perspective variations of crowd in the patch.

We use three CNN regressors introduced in \cite{zhang2016single}, $R_1$ through $R_3$, in Switch-CNN to predict the density of crowd. These CNN regressors have varying receptive fields that can capture people at
different scales. The architecture of each of the shallow CNN regressor is similar: four convolutional layers
with two pooling layers. $R_1$ has a large initial filter
size of 9$\times$9 which can capture high level abstractions within the scene like faces, urban facade etc. $R_2$ and $R_3$ with initial filter sizes 7$\times$7 and 5$\times$5 capture crowds at lower scales detecting blob like abstractions. 

Patches are relayed to a regressor using a switch. The switch consists of a switch classifier and a switch layer. The switch classifier infers the label of the regressor to which the patch is to be relayed to. A switch layer takes the label inferred from the switch classifier and relays it to the correct regressor. For example, in Figure~\ref{arch}, the switch classifier relays the patch highlighted in \textcolor{red}{red} to regressor $R_3$. The patch has a very high crowd density. Switch relays it to regressor $R_3$ which has smaller receptive field: ideal for detecting blob like abstractions characteristic of patches with high crowd density. We use an adaptation of VGG16~\cite{simonyan2014very} network as the switch classifier to perform 3-way classification. The fully-connected layers in VGG16 are removed.  We use global average pool (GAP) on Conv5 features to remove the spatial information and aggregate discriminative features. GAP is
followed by a smaller fully connected layer and 3-class softmax classifier corresponding to the three regressor networks in Switch-CNN. 

\textbf{Ground Truth} Annotations for crowd images are provided as point annotations at the center of the head of a person. We generate our ground truth by blurring each head annotation with a Gaussian kernel normalized to sum to one to generate a density map. Summing the resultant density map gives the crowd count. Density maps ease the difficulty of regression for the CNN as the task of predicting the exact point of head annotation is reduced to predicting a coarse location. The spread of the Gaussian in the above density map is fixed. However, a density map generated from a fixed spread Gaussian is inappropriate if the variation in crowd density is large. We use geometry-adaptive kernels\cite{zhang2016single} to vary the spread parameter
of the Gaussian depending on the local crowd density. It sets the spread of Gaussian in proportion to the average distance of $k$-nearest neighboring head
annotations. The inter-head distance is a good substitute for perspective maps which are laborious to generate and unavailable for every dataset. This results in lower degree of Gaussian blur for dense crowds and higher degree for region of sparse density in crowd scene. In our experiments, we use both geometry-adaptive kernel method as well as fixed spread Gaussian method to generate
ground truth density depending on the dataset. Geometry-adaptive
kernel method is used to generate ground truth density maps for datasets with dense crowds and large variation in count across scenes. Datasets that have sparse crowds are trained using density maps generated from fixed spread Gaussian method.

Training of Switch-CNN is done in three stages, namely pretraining, differential training and coupled training described in Sec \ref{a1}--\ref{a4}.

\subsection{Pretraining}
\label{a1}
The three CNN regressors $R_1$ through $R_3$ are pretrained separately to regress density maps. Pretraining helps in learning good initial features which improves later fine-tuning stages. Individual CNN regressors are trained to minimize the Euclidean distance
between the estimated density map and ground truth. Let $D_{X_{i}}(\bigcdot;\Theta)$
represent the output of a CNN regressor with parameters $\Theta$
for an input image $X_{i}$. The $l_2$ loss function is given by 
\begin{equation}
L_{l_2}(\Theta)=\frac{1}{2N}\sum_{i=1}^{N}\|D_{X_{i}}(\bigcdot;\Theta)-D_{X_{i}}^{GT}(\bigcdot)\|_{2}^{2},
\label{eq1}
\end{equation}

where $N$ is the number of training samples and $D_{X_{i}}^{GT}(\bigcdot)$
indicates ground truth density map for image $X_{i}$. The loss $L_{l_2}$
is optimized by backpropagating the CNN via stochastic gradient descent
(SGD). Here, $l_2$ loss function acts as a proxy for count error 
between the regressor estimated count and true count. It indirectly minimizes count error.
The regressors $R_k$ are pretrained until the validation accuracy plateaus.

\subsection{Differential Training}

CNN regressors $R_{1-3}$ are pretrained with the entire
training data. The count prediction performance varies due to the inherent difference
in network structure of  $R_{1-3}$ like receptive field and effective field-of-view. 
Though we optimize the $l_2$-loss between the estimated and ground
truth density maps for training CNN regressor, factoring in count error during training leads to better crowd counting performance. Hence, we measure CNN performance using count error. Let
the count estimated by $k$th regressor for $i$th image be $C_{i}^{k}=\sum_{x}D_{X_{i}}(x;\Theta_{k})$
. Let the reference count inferred from ground truth be $C_{i}^{GT}=\sum_{x}D_{X_{i}}^{GT}(x)$. Then
count error for $i$th sample evaluated by $R_k$ is
\begin{equation}
E_{C_{i}}(k)=|C_{i}^{k}-C_{i}^{GT}|,
\label{eq2}
\end{equation}
the absolute count difference between prediction and true count. Patches with particular crowd attributes give lower count error with a regressor having complementary network structure. For example, a CNN regressor with large receptive field capture high level abstractions like background elements and faces. To amplify
the network differences, differential training is proposed (shown in \textcolor{BlueViolet}{blue} in Algorithm \ref{algo}). The key
idea in differential training is to backpropagate the regressor $R_k$ with minimum count error for a given training crowd scene patch. For every training patch $i$, we choose the
regressor $l_{i}^{best}$ such that $E_{C_{i}}(l_{i}^{best})$ is lowest across all regressors $R_{1-3}$. This amounts to greedily choosing the regressor that predicts the most accurate count amongst $k$ regressors.  Formally, we define the label of chosen regressor $l_{i}^{best}$ as:

\begin{equation}
l_{i}^{best}=\underset{k}{\textrm{argmin}}|C_{i}^{k}-C_{i}^{GT}|
\label{eqm}
\end{equation}

The count error for $i$th sample is 
\begin{equation}
E_{C_{i}}=\underset{k}{\textrm{min}}|C_{i}^{k}-C_{i}^{GT}|.
\label{eq3}
\end{equation}
This training regime encourages a regressor $R_k$ to prefer a particular set of the
training data patches with particular patch attribute so as to minimize the loss. While the backpropagation of independent regressor $R_k$ is still done with $l_2$-loss,
 the choice of CNN regressor for backpropagation is based on the count error. Differential training indirectly minimizes the mean absolute count error (MAE) over the training images.
For $N$ images, MAE in this case is given by 
\begin{equation}
E_{C}=\frac{1}{N}\sum_{i=1}^{N}\underset{k}{\textrm{min}}|C_{i}^{k}-C_{i}^{GT}|,
\label{eq4}
\end{equation}
which can be thought as the minimum count error achievable if each
sample is relayed correctly to the right CNN. However during testing,
achieving this full accuracy may not be possible as the switch classifier is not ideal. To summarize, differential training generates three disjoint groups of training patches and each network is finetuned on its own group. The regressors $R_k$ are differentially trained until the validation accuracy plateaus.

\subsection{Switch Training}

Once the multichotomy of space of patches is inferred via differential training, a patch classifier (switch) is trained to relay a patch to the correct regressor $R_k$. The manifold that separates the space of crowd scene patches is complex and hence a deep classifier is required to infer the group of patches in the multichotomy. We use VGG16~\cite{simonyan2014very} network as the switch classifier to perform 3-way classification. The classifier is trained on the labels of multichotomy generated from differential training. The number of training patches in each group can be highly skewed, with the majority of patches being relayed to a single regressor depending on the attributes of crowd scene. To alleviate class imbalance during switch classifier training, the labels collected from the differential training are equalized so
that the number of samples in each group is the same. This is done by
randomly sampling from the smaller group to balance the training set of switch classifier.

\subsection{Coupled Training}
\label{a4}
Differential training on the CNN regressors $R_1$ through $R_3$ generates a multichotomy that minimizes the predicted count by choosing the best regressor for a given crowd scene patch. However, the trained switch is not ideal and the manifold separating the space of patches is complex to learn. To mitigate the effect of switch inaccuracy and inherent complexity of task, we co-adapt the patch classifier and the CNN regressors by training the switch and regressors in an alternating fashion. We refer to this stage of training as \emph{Coupled training} (shown in \textcolor{OliveGreen}{green} in Algorithm \ref{algo}). 

The switch classifier is first trained
with labels from the multichotomy inferred in differential training for one epoch (shown in \textcolor{red}{red} in Algorithm \ref{algo}). In, the next stage, the three CNN regressors are made to co-adapt with switch classifier (shown in \textcolor{BlueViolet}{blue} in Algorithm \ref{algo}). We refer to this stage of training enforcing co-adaption of switch and regressor $R_{1-3}$ as \emph{Switched differential training}.

In switched differential training, the individual CNN regressors are trained using crowd scene patches relayed by switch for one epoch. For a given training crowd scene patch $X_i$, switch is forward propagated on $X_i$ to infer the choice of regressor $R_k$. The switch layer then relays $X_i$ to the particular regressor and backpropagates $R_k$ using the loss defined in Equation \ref{eq1} and $\theta_k$ is updated. This training regime is executed for an epoch. 

In the next epoch, the labels for training the switch classifier are recomputed using criterion in Equation \ref{eqm} and the switch is again trained as described above. This process of alternating switch training and switched training of CNN regressors is repeated every epoch until the validation accuracy plateaus.



\section{Experiments}

\subsection{Testing} We evaluate the performance of our proposed architecture, Switch-CNN on four major crowd counting datasets 
At test time, the image patches are fed to the switch classifier which relays the patch to the best CNN regressor $R_k$. The selected CNN regressor predicts a crowd density map for the relayed crowd scene patch. The generated density maps are assembled into an image to get the final density map for the entire scene. Because of the two pooling layers in the CNN regressors, the predicted density maps are $\frac{1}{4}$\textsuperscript{th} size of the input. \newline\\
\textbf{Evaluation Metric} We use Mean Absolute Error (MAE) and Mean Squared Error
(MSE) as the metric for comparing the performance of Switch-CNN against the state-of-the-art crowd counting methods. For a test sequence with $N$ images, MAE is defined as follows:
\begin{equation}
\textrm{MAE}=\frac{1}{N}\sum_{i=1}^{N}|C_{i}-C_{i}^{GT}|,
\label{mae}
\end{equation}
where $C_{i}$ is the crowd count predicted by the model being evaluated, and $C_{i}^{GT}$ is the
crowd count from human labelled annotations. MAE is an indicator of the accuracy of the predicted crowd count across the test sequence. MSE is a metric complementary to MAE and indicates the robustness of the predicted count. For a test sequence, MSE is defined as follows:
\begin{equation}
\textrm{MSE}=\sqrt{\frac{1}{N}\sum_{i=1}^{N}(C_{i}-C_{i}^{GT})^{2}}.
\label{mse}
\end{equation}

\subsection{ShanghaiTech dataset}

We perform extensive experiments on the ShanghaiTech crowd counting dataset~\cite{zhang2016single} that consists of 1198 annotated images. The dataset is divided into two parts named Part A and Part B. The former contains dense crowd scenes parsed from the internet and the latter is relatively sparse crowd scenes captured in urban surface streets. We use the train-test splits provided by the authors for both parts in our experiments. We train Switch-CNN as elucidated by Algorithm \ref{algo} on both parts of the dataset. Ground truth is generated
using geometry-adaptive kernels method as the variance in crowd density within a scene due to perspective effects is high (See Sec \ref{scnn} for details about ground truth generation). With an ideal switch (100\% switching accuracy), Switch-CNN performs with an MAE of 51.4. However, the accuracy of the switch is 73.2\% in Part A and 76.3\% in Part B of the dataset resulting in a lower MAE. 

Table \ref{shanghaitech} shows that Switch-CNN outperforms all other state-of-the art methods by a significant margin on both the MAE and MSE metric. Switch-CNN shows a 19.8 point improvement in MAE on Part A and 4.8 point improvement in Part B of the dataset over  MCNN~\cite{zhang2016single}. Switch-CNN also outperforms all other models on MSE metric indicating that the predictions have a lower variance than MCNN across the dataset. This is an indicator of the robustness of Switch-CNN's predicted crowd count.

\begin{figure}[!t]
 \includegraphics[width =0.48\textwidth, height=4.0cm]{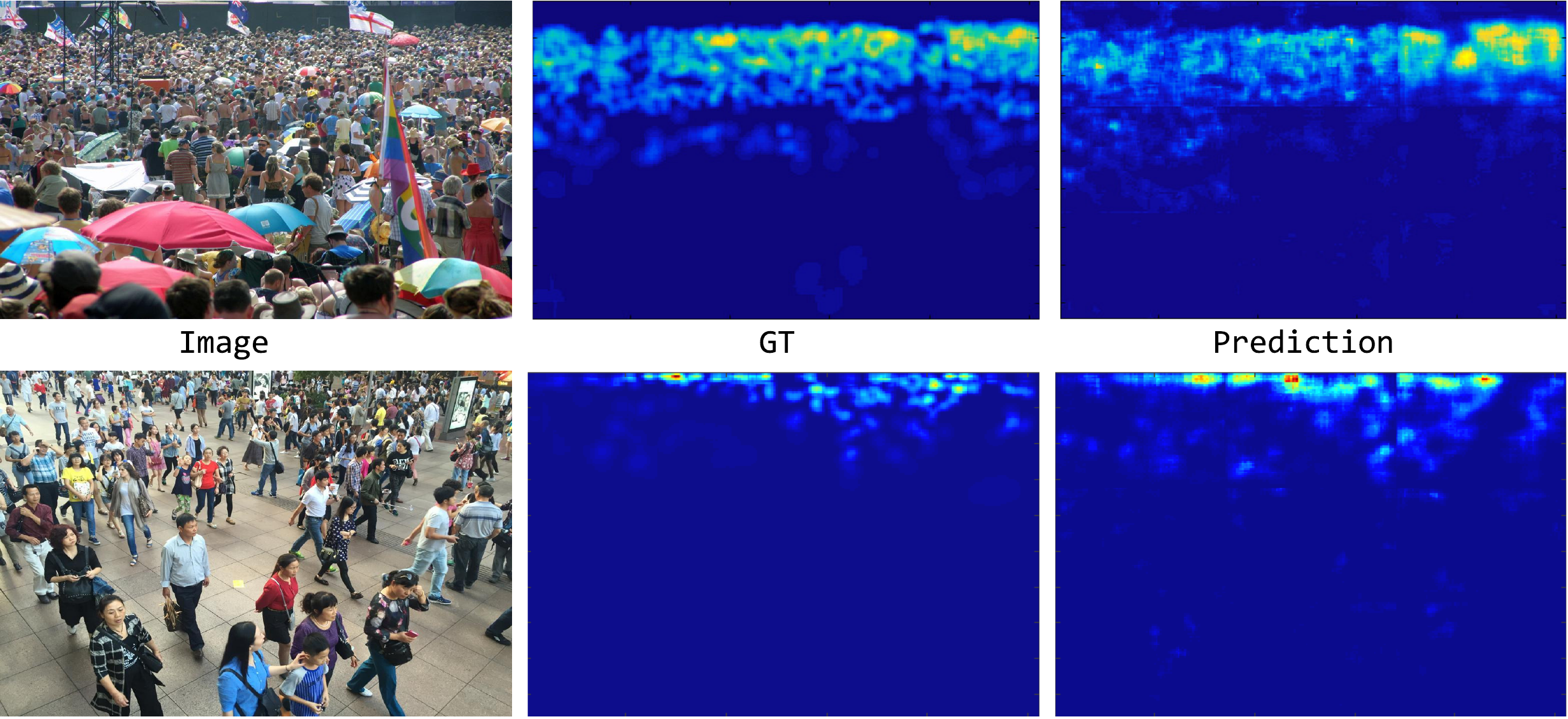}\hspace{0.3cm}%
 \caption{Sample predictions by Switch-CNN for crowd scenes from the ShanghaiTech dataset~\cite{zhang2016single} is shown. The top and bottom rows depict a crowd image, corresponding ground truth and prediction from Part A and Part B of dataset respectively.}
\label{stpred}
\end{figure}

We show sample predictions of Switch-CNN for sample test scenes from the ShanghaiTech dataset along with the ground truth in Figure~\ref{stpred}. The predicted density maps closely follow the crowd distribution visually. This indicates that Switch-CNN is able to localize the spatial distribution of crowd within a scene accurately.

\begin{table}[!h]
\centering
\scalebox{0.8}{
\begin{tabular}{|c|c|c|c|c|}
\hline 
 & \multicolumn{2}{c|}{Part A} & \multicolumn{2}{c|}{Part B}\tabularnewline
\hline 
Method & MAE & MSE & MAE & MSE\tabularnewline
\hline 
\hline 
Zhang et al.~\cite{zhang2015cross} & 181.8 & 277.7 & 32.0 & 49.8\tabularnewline
\hline 
MCNN~\cite{zhang2016single} & 110.2 & 173.2 & 26.4 & 41.3\tabularnewline
\hline
\textbf{Switch-CNN} & \textbf{90.4} & \textbf{135.0} & \textbf{21.6} & \textbf{33.4} \tabularnewline
\hline 
\end{tabular}\medskip{}}

\vspace*{1.2mm}
\caption{Comparison of Switch-CNN with other state-of-the-art crowd counting methods on ShanghaiTech dataset~\cite{zhang2016single}.}
\label{shanghaitech}

\end{table}

\subsection{UCF\_CC\_50 dataset}

UCF\_CC\_50~\cite{idrees2013multi} is a 50 image collection of annotated crowd scenes. The dataset exhibits a large variance in the crowd count  with counts varying between
94 and 4543. The small size of the dataset and large variance in crowd count makes it a very challenging dataset. We follow the approach of other state-of-the-art models~\cite{zhang2015cross,boominathan2016crowdnet,onoro2016towards,zhang2016single} and use 5-fold cross-validation to validate the performance of Switch-CNN on UCF\_CC\_50.  

In Table \ref{ucf50}, we compare the performance of Switch-CNN with other methods using MAE and MSE as metrics. Switch-CNN outperforms all other methods and evidences a 15.7 point improvement in MAE over Hydra2s~\cite{onoro2016towards}. Switch-CNN also gets a competitive MSE score compared to Hydra2s indicating the robustness of the predicted count. The accuracy of the switch  is 54.3\%. The switch accuracy is relatively low as the dataset has very few training examples and a large variation in crowd density. This limits the ability of the switch to learn the multichotomy of space of crowd scene patches.

\begin{table}[!h]
\centering
\scalebox{0.8}{
\begin{tabular}{|c|c|c|}
\hline 
Method & MAE & MSE\tabularnewline
\hline 
\hline 
Lempitsky et al.\cite{lempitsky2010learning} & 493.4 & 487.1\tabularnewline
\hline 
Idrees et al.\cite{idrees2013multi} & 419.5 & 487.1\tabularnewline
\hline 
Zhang et al.~\cite{zhang2015cross}  & 467.0 & 498.5\tabularnewline
\hline 
CrowdNet~\cite{boominathan2016crowdnet} & 452.5 & -- \tabularnewline
\hline 
MCNN~\cite{zhang2016single} & 377.6 & 509.1\tabularnewline
\hline 
Hydra2s~\cite{onoro2016towards} & 333.73  & \textbf{425.26} \tabularnewline
\hline 
\textbf{Switch-CNN} & \textbf{318.1} & 439.2 \tabularnewline
\hline 
\end{tabular}\medskip{}}

\vspace*{1.2mm}
\caption{Comparison of Switch-CNN with other state-of-the-art crowd counting methods on UCF\_CC\_50 dataset~\cite{idrees2013multi}.}
\label{ucf50}
\end{table}

\subsection{The UCSD dataset}

The UCSD dataset crowd counting dataset consists of 2000 frames from a single scene. The scenes are characterized by sparse crowd with the number of people ranging from 11 to 46 per frame. A region of interest (ROI) is provided for the scene in the dataset. We use the train-test splits used by \cite{chan2008privacy}. Of the 2000 frames, frames 601 through 1400 are used for training while the remaining frames are held out for testing. Following the setting used in \cite{zhang2016single}, we prune the feature maps of the last layer with the ROI provided. Hence, error is backpropagated during training for areas inside the ROI. We use a fixed spread Gaussian to generate ground truth density maps for training Switch-CNN as the crowd is relatively sparse. At test time, MAE is computed only for the specified ROI in test images for benchmarking Switch-CNN against other approaches.

Table \ref{ucsd} reports the MAE and MSE results for Switch-CNN and other state-of-the-art approaches. Switch-CNN performs competitively compared to other approaches with an MAE of 1.62. The switch accuracy in relaying the patches to regressors $R_1$ through $R_3$ is 60.9\%. However, the dataset is characterized by low variability of crowd density set in a single scene. This limits the performance gain achieved by Switch-CNN from leveraging intra-scene crowd density variation.

\begin{table}[!h]
\centering
\scalebox{0.9}{
\begin{tabular}{|c|c|c|}
\hline 
Method & MAE & MSE\tabularnewline
\hline 
\hline 
Kernel Ridge Regression~\cite{an2007face}& 2.16 & 7.45\tabularnewline
\hline 
Cumulative Attribute Regression~\cite{chen2012feature} & 2.07 & 6.86\tabularnewline
\hline 
Zhang et al.~\cite{zhang2015cross} & 1.60 & 3.31\tabularnewline
\hline 
MCNN~\cite{zhang2016single} & \textbf{1.07} & \textbf{1.35}\tabularnewline
\hline 
CCNN~\cite{onoro2016towards} & 1.51  & -- \tabularnewline
\hline 
\textbf{Switch-CNN} & 1.62  & 2.10 \tabularnewline
\hline 
\end{tabular}\medskip{}}

\vspace*{1.2mm}
\caption{Comparison of Switch-CNN with other state-of-the-art crowd counting methods on UCSD crowd-counting  dataset~\cite{chan2008privacy}.}
\label{ucsd}
\end{table}

\subsection{The WorldExpo'10 dataset}

\begin{table}[!t]
\centering
\vspace{-0.2cm}
\scalebox{0.8}{
\begin{tabular}{|c|c|c|c|c|c|c|}
\hline 
Method & S1 & S2 & S3 & S4 & S5 & Avg.\tabularnewline
& &  & &  &  & MAE\tabularnewline
\hline 
\hline 
Zhang et al.~\cite{zhang2015cross} & 9.8 & \textbf{14.1} & 14.3 & 22.2 & \textbf{3.7} & 12.9\tabularnewline
\hline 
MCNN~\cite{zhang2016single} & \textbf{3.4} & 20.6 & 12.9 & 13.0 & 8.1 & 11.6\tabularnewline
\hline
Switch-CNN & 4.2  & 14.9 & 14.2 & 18.7 & 4.3 & 11.2 \tabularnewline
(GT with perspective map) & & & & & &\tabularnewline
\hline 
\textbf{Switch-CNN} & 4.4 & 15.7 & \textbf{10.0} & \textbf{11.0} & 5.9 & \textbf{9.4} \tabularnewline
\textbf{(GT without perspective)} & & & & & &\tabularnewline
\hline 
\end{tabular}\medskip{}}

\vspace*{1.2mm}
\caption{Comparison of Switch-CNN with other state-of-the-art crowd counting methods on WorldExpo'10 dataset~\cite{zhang2015cross}. Mean Absolute Error (MAE) for individual test scenes and average performance across scenes is shown.}
\label{wexpo}
\end{table}

The WorldExpo'10 dateset consists of 1132 video sequences captured with 108 surveillance cameras. Five different video sequence, each from a different scene, are held out for testing. Every test scene sequence has 120 frames. The crowds are relatively sparse in comparison to other datasets with average number of 50 people per image. Region of interest (ROI) is provided for both training and test scenes. In addition, perspective maps are provided for all scenes. The maps specify the number of pixels in the image that cover one square meter at every location in the frame. These maps are used by \cite{zhang2016single,zhang2015cross} to adaptively choose the spread of the Gaussian while generating ground truth density maps. We evaluate performance of the Switch-CNN using ground truth generated with and without perspective maps. 

We prune the feature maps of the last layer with the ROI provided. Hence, error is backpropagated during training for areas inside the ROI. Similarly at test time, MAE is computed only for the specified ROI in test images for benchmarking Switch-CNN against other approaches.

MAE is computed separately for each test scene and averaged to determine the overall performance of Switch-CNN across test scenes. Table \ref{wexpo} shows that the average MAE of Switch-CNN across scenes is better by a margin of 2.2 point over the performance obtained by the state-of-the-art approach MCNN~\cite{zhang2016single}. The switch accuracy is 52.72\%. 



\section{Analysis}

\subsection{Effect of number of regressors on Switch-CNN}
Differential training makes use of the structural variations across
the individual regressors to learn a multichotomy of the training data. To investigate the effect of structural variations of the regressors $R_1$ through $R_3$, we train Switch-CNN with combinations of regressors ($R_1$,$R_2$), ($R_2$,$R_3$), ($R_1$,$R_3$) and ($R_1$,$R_2$,$R_3$) on Part A of ShanghaiTech dataset. Table \ref{comp2} shows the MAE performance of Switch-CNN for different combinations of regressors $R_k$. Switch-CNN with CNN regressors $R_1$ and $R_3$ has lower MAE
than Switch-CNN with regressors $R_1$--$R_2$ and $R_2$--$R_3$. This can be attributed to the former model having a higher switching accuracy than the latter. Switch-CNN with all three regressors outperforms both the models as it is able to model the scale and perspective variations better with three independent CNN regressors $R_1$, $R_2$ and $R_3$ that are structurally distinct.
Switch-CNN leverages multiple independent CNN regressors with different receptive fields. In Table \ref{comp2}, we also compare the performance of individual CNN regressors with Switch-CNN. Here each of the individual regressors are trained on the full training data from Part A of Shanghaitech dataset. The higher MAE of the individual CNN regressor is attributed to the inability of a single regressor to model the scale and perspective variations in the crowd scene. 

\begin{table}[tbh]
\begin{centering}
\scalebox{0.8}{
\begin{tabular}{|c|c|}
\hline 
Method & MAE\tabularnewline
\hline 
\hline 
$R_1$  & 157.61\tabularnewline
\hline 
$R_2$ & 178.82\tabularnewline
\hline 
$R_3$ & 178.10\tabularnewline
\hline
Switch-CNN with ($R_1$,$R_3$)  & 98.87\tabularnewline
\hline 
Switch-CNN with ($R_1$,$R_2$)  & 110.88\tabularnewline
\hline
Switch-CNN with ($R_2$,$R_3$)  & 126.65\tabularnewline
\hline
\textbf{Switch-CNN} with ($R_1$,$R_2$,$R_3$) & \textbf{90.41}\tabularnewline
\hline 
\end{tabular}\medskip{}}

\par\end{centering}
\vspace{2mm}
\caption{Comparison of MAE for Switch-CNN variants and CNN regressors $R_1$ through $R_3$ on Part A of the ShanghaiTech dataset~\cite{zhang2016single}.}
\label{comp2}
\end{table}

\subsection{Switch Multichotomy Characteristics}
\label{char}

The principal idea of Switch-CNN is to divide the training patches into disjoint groups to train individual CNN regressors so that overall count accuracy is maximized.
This multichotomy in space of crowd scene patches is created automatically through differential training.
We examine the underlying structure of the patches to understand the correlation between the learnt multichotomy and attributes of the patch like crowd count and density. However, the unavailability of perspective maps renders computation of actual density intractable. We believe inter-head distance between people is a candidate measure of crowd density. In a highly dense crowd, the separation between people is
low and hence density is high. On the other hand, for low density
scenes, people are far away and mean inter-head distance is large.
Thus mean inter-head distance is a proxy for crowd density. This measure of density is robust to scale variations as the inter-head distance naturally subsumes the scale variations. 

\begin{figure}[!h]
 \includegraphics[width =0.5\textwidth, height=4.5cm]{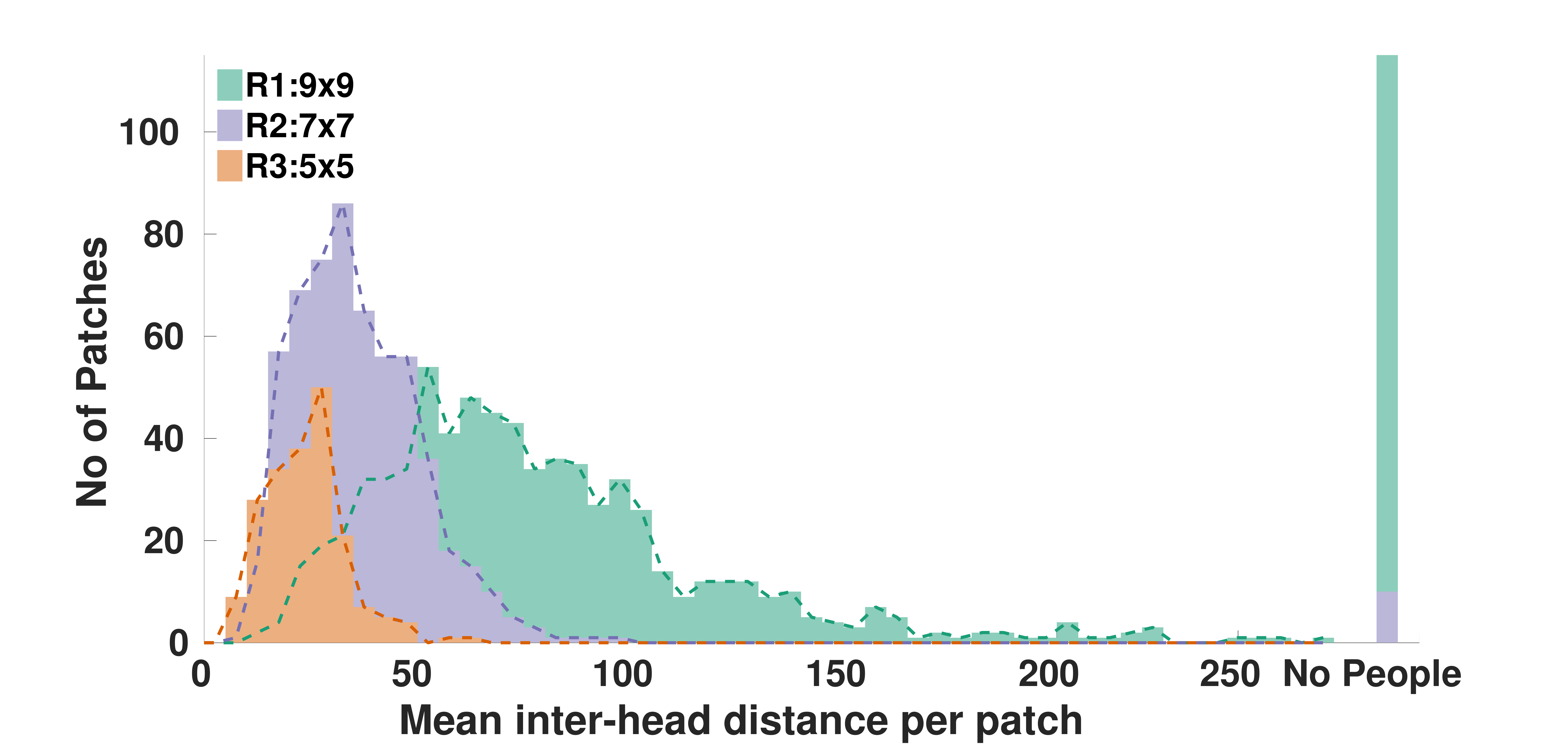}\hspace{0.3cm}%
 \caption{Histogram of average inter-head distance for crowd scene patches from Part A test set of ShanghaiTech dataset~\cite{zhang2016single} is shown in Figure~\ref{cluster_plot}. We see that the multichotomy of space of crowd scene patches inferred from the switch separates patches based on latent factors correlated with crowd density.}
\label{cluster_plot}
\end{figure}

\begin{figure}[!t]
   \vspace{-.3cm}
 \includegraphics[width =0.48\textwidth, height=4.0cm]{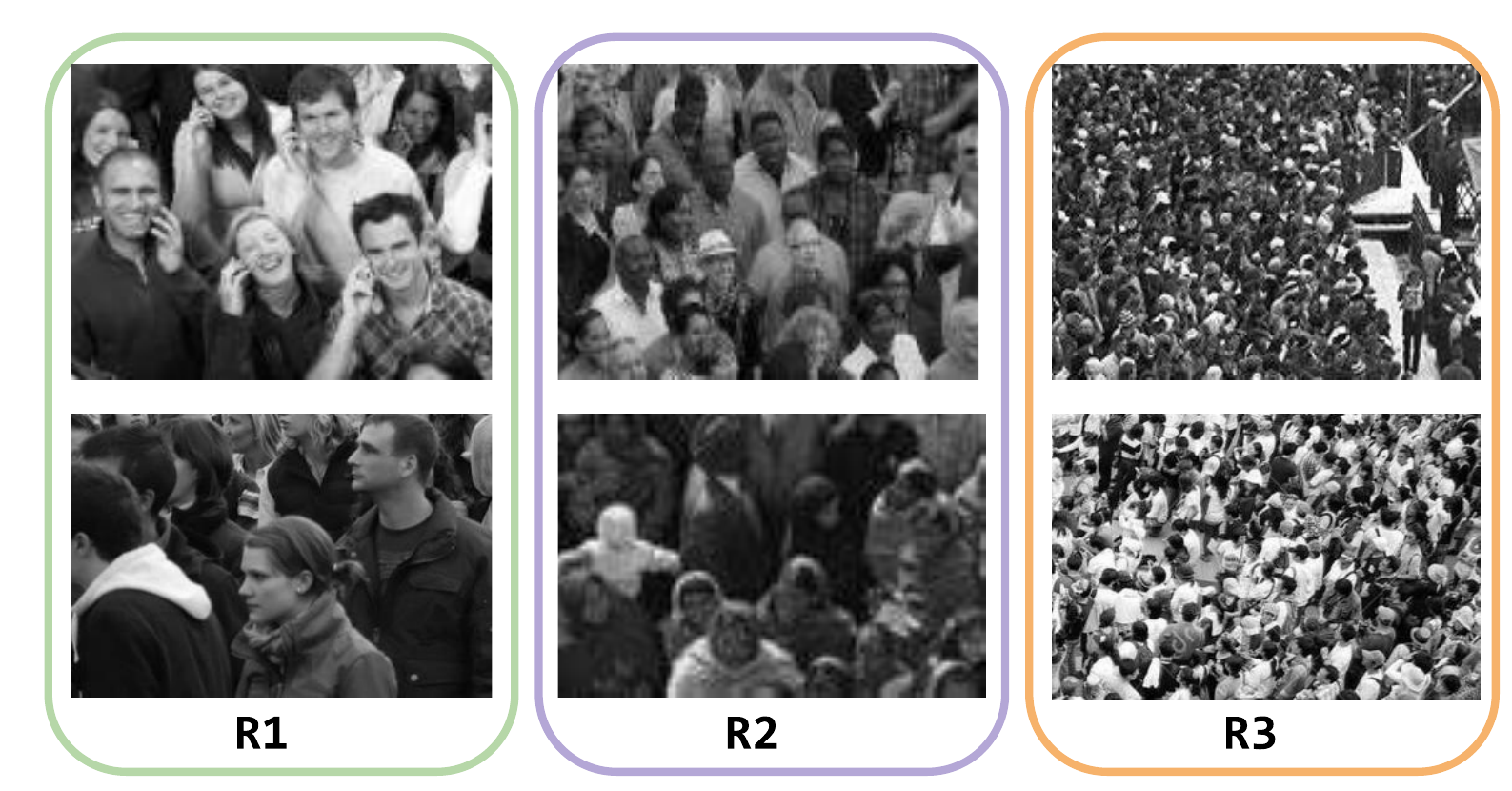}\hspace{0.3cm}%
 \caption{Sample crowd scene patches from Part A test set of ShanghaiTech dataset~\cite{zhang2016single} are shown. We see that the density of crowd in the patches increases from CNN regressor $R_1$--$R_3$.}
\label{mosaic_switch}
\end{figure}

To analyze the multichotomy in space of patches, we compute the average inter-head distance of each patch in Part A of ShanghaiTech test set. For each head annotation, the average distance to its 10 nearest neighbors is calculated.
These distances are averaged over the entire patch representing the density of the patch. We plot a histogram of these distances in Figure~\ref{cluster_plot} and group the patches by color on the basis of the regressor $R_k$ used to infer the count of the patch. A separation of patch space based on crowd density is observed in Figure~\ref{cluster_plot}. $R_1$, which has the largest receptive
field of 9$\times$9, evaluates patches of low crowd density (corresponding to large mean inter-head distance).  An interesting observation is that patches from the crowd scene that have no people in them (patches in Figure~\ref{cluster_plot} with zero average inter-head distance) are relayed to $R_1$ by the switch. We believe that the patches with no people are relayed to $R_1$ as it has a large receptive field that helps capture background attributes in such patches like urban facade and foliage. Figure~\ref{mosaic_switch} displays some sample patches that are relayed to each of the CNN regressors $R_1$ through $R_3$. The density of crowd in the patches increases from CNN regressor $R_1$ through $R_3$.

\subsection{Attribute Clustering Vs Differential Training}

We saw in Sec \ref{char} that differential training approximately divides
training set patches into a multichotomy based on density. We investigate the effect of manually clustering the patches based on patch attribute like crowd count or density. We use patch count as metric to cluster patches.
Training patches are divided into three groups based on the patch
count such that the total number of training patches are equally distributed amongst the three CNN regressors $R_{1-3}$.
$R_1$, having a large receptive field, is trained on patches with low
crowd count. $R_2$ is trained on medium count patches while high count
patches are relayed to $R_3$. The training procedure for this experiment is identical to Switch-CNN, except for the differential training stage. We repeat this 
experiment with average inter-head distance of the patches as a metric for grouping the patches. Patches with high mean inter-head distance are relayed
to $R_1$. $R_2$ is relayed patches with low inter-head distance by the switch
while the remaining patches are relayed to $R_3$. 
\begin{table}[!h]
\begin{centering}
\scalebox{0.8}{
\begin{tabular}{|c|c|}
\hline 
Method & MAE\tabularnewline
\hline 
\hline 
Cluster by count  & 99.56\tabularnewline
\hline 
Cluster by mean inter-head distance  & 94.93\tabularnewline
\hline 
\textbf{Switch-CNN} & \textbf{90.41}\tabularnewline
\hline 
\end{tabular}\medskip{}}

\par\end{centering}
\vspace*{1.2mm}
\caption{Comparison of MAE for Switch-CNN and manual clustering of patches based on patch attributes on Part A of the ShanghaiTech dataset~\cite{zhang2016single}.}
\label{man}
\end{table}

Table \ref{man} reports  MAE performance for the two clustering methods. Both crowd count and average inter-head distance based clustering give a higher MAE than Switch-CNN. Average inter-head distance based clustering performs comparably with Switch-CNN. This evidence reinforces the fact that Switch-CNN  learns a multichotomy in the space of patches that is highly correlated with mean inter-head distance of the
crowd scene. The differential training regime employed by Switch-CNN is able to infer this grouping automatically, independent of the dataset.


\subsection{Effect of Coupled Training}
\label{a1}
Differential training on the CNN regressors $R_1$ through $R_3$ generates a multichotomy that minimizes the predicted count by choosing the best regressor for a given crowd scene patch. However, the trained switch is not ideal and the manifold separating the space of patches is complex to learn (see Section 5.2 of the main paper). To mitigate the effect of switch inaccuracy and inherent complexity of task, we perform coupled training of switch and CNN regressors. We ablate the effect of coupled training by training the switch classifier in a stand-alone fashion. For training the switch in a stand-alone fashion, the labels from differential training are held fixed throughout the switch classifier training.

The results of the ablation are reported in Table \ref{t1}. We see that training the switch classifier in a stand-alone fashion results in a deterioration of Switch-CNN crowd counting performance. While Switch-CNN with the switch trained in a stand-alone manner performs better than MCNN, it performs significantly worse than Switch-CNN with coupled training. This is reflected in the 13 point higher count MAE. Coupled training allows the patch labels to change in order to adapt to the ability of the switch classifier to relay a patch to the optimal regressor $R_k$ correctly. This co-adaption is absent when training switch alone leading to deterioration of crowd counting performance.
\begin{table}[!h]
\begin{centering}
\scalebox{0.8}{
\begin{tabular}{|c|c|}
\hline 
Method & MAE\tabularnewline
\hline 
\hline 
MCNN~\cite{zhang2016single} & 110.2 \tabularnewline
\hline
Switch-CNN without Coupled Training & 103.26\tabularnewline
\hline 
\textbf{Switch-CNN with Coupled Training} & \textbf{90.41}\tabularnewline
\hline 
\end{tabular}\medskip{}}

\par\end{centering}
\vspace*{1.2mm}
\caption{Comparison of MAE for Switch-CNN trained with and without \emph{Coupled Training} on Part A of the ShanghaiTech dataset~\cite{zhang2016single}.}
\label{t1}
\end{table}

\subsection{Ablations on UCF\_CC\_50 dataset}
 We perform ablations referenced in Section 5.1 and 5.3 of the main paper on the UCF\_CC\_50 dataset~\cite{idrees2013multi}. The results of these ablations are tabulated in Table~\ref{tend}. The results follow the trend on ShanghaiTech dataset and reinforce the superiority of Switch-CNN (See Section 5.1 and 5.3 of the main paper for more details).

\begin{table}[!h]
\centering

\vspace{.1cm}
\scalebox{0.8}{
\begin{tabular}{|c|c|}
\hline 
Method & MAE \tabularnewline
\hline 
\hline 
Cluster by count  & 319.16 \tabularnewline
\hline
Cluster by mean inter-head distance & 358.78 \tabularnewline
\hline
Switch-CNN($R_1$,$R_3$)  & 369.58 \tabularnewline
\hline
Switch-CNN($R_1$,$R_2$)  & 362.22 \tabularnewline
\hline 
Switch-CNN($R_3$,$R_2$) & 334.66\tabularnewline
\hline
\textbf{Switch-CNN} & \textbf{318.07}\tabularnewline
\hline 
\end{tabular}}\medskip{}
\caption{Additional results for ablations referenced in Section 5.1 and 5.3 of the main paper for UCF\_CC\_50 dataset\cite{idrees2013multi}.}
\label{tend}
\end{table}

\subsection{Choice of Switch Classifier}
The switch classifier is used to infer the multichotomy of crowd patches learnt from differential training. The accuracy of the predicted count in Switch-CNN is critically dependent on the choice of the switch classifier. We repurpose different classifier architectures, from shallow CNN classifiers to state-of-the art object classifiers to choose the best classifier that strikes a balance between classification accuracy and computational complexity.

\begin{table}[!t]
\begin{centering}
\scalebox{0.8}{
\begin{tabular}{|c|c|}
\hline 
Method & Acc\tabularnewline
\hline 
\hline 
CNN-small  & 64.39 \tabularnewline
\hline 
\textcolor{red}{VGG-16}  & \textcolor{red}{73.75}\tabularnewline
\hline 
VGG-19  & 74.3\tabularnewline
\hline 
ResNet-50  & 75.03\tabularnewline
\hline 
ResNet-101  & 74.95 \tabularnewline
\hline 
\end{tabular}\medskip{}}

\par\end{centering}
\vspace*{1.2mm}
\caption{Comparison of classification accuracy for different switch architectures on Part A of the ShanghaiTech dataset~\cite{zhang2016single}. The final switch-classifier selected for all Switch-CNN experiments is highlighted in \textcolor{red}{red}. }
\label{t2}
\end{table}

\begin{figure*}[t]
\centering
 \includegraphics[width=17.0cm]{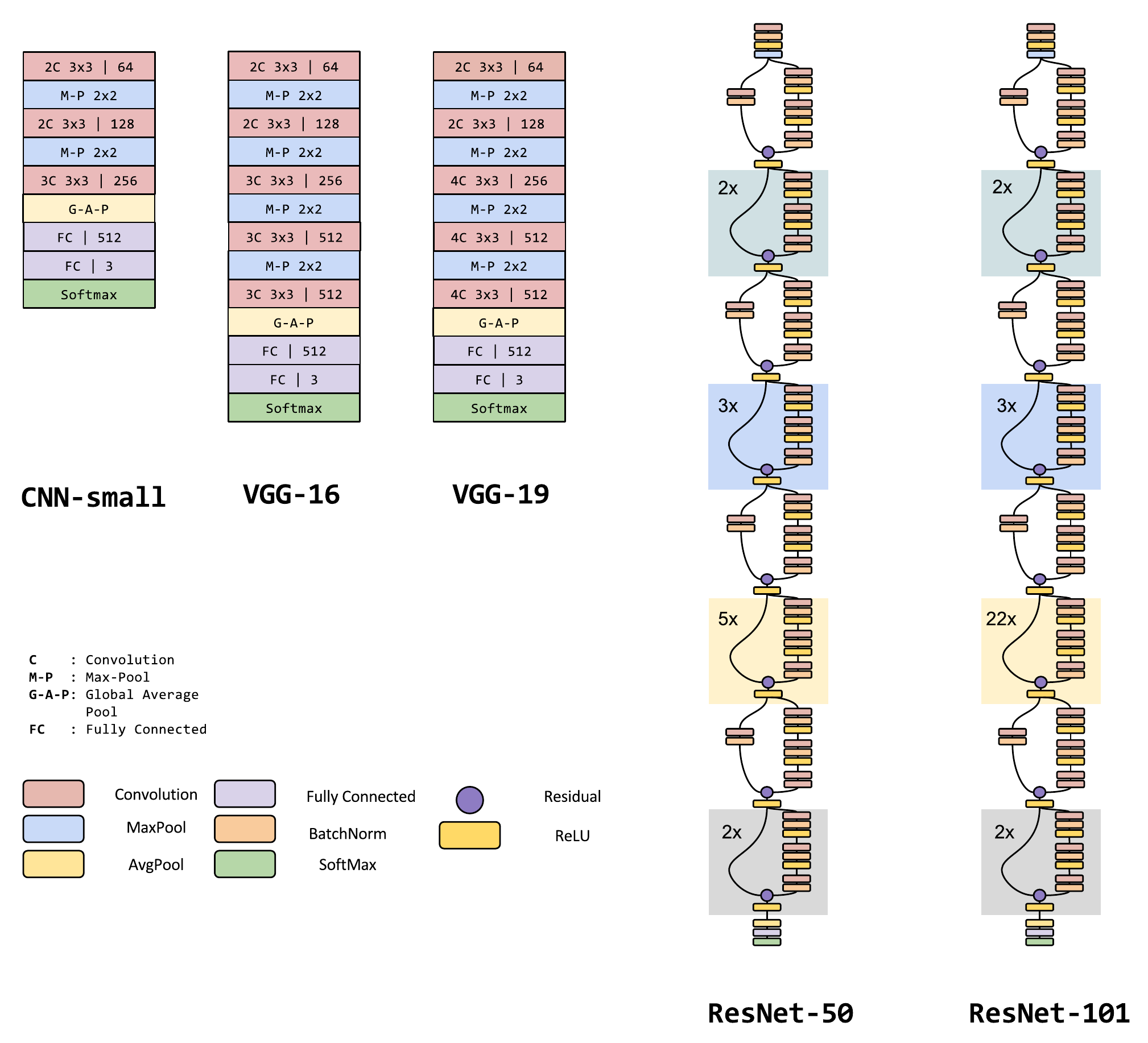}\hspace{0.3cm}%
 \caption{The architecture of different switch classifiers evaluated in \textbf{Switch-CNN}.}
\label{switch_list}
\end{figure*}
Figure~\ref{switch_list} shows the different architectures of switch classifier that we evaluate. CNN-small is a shallow classifier derived from VGG-16~\cite{simonyan2014very}. We retain the first three convolutional layers from VGG-16 and add a 512 dimensional fully-connected layer along with a 3-way classifier. The convolutional layers in CNN-small are initialized from VGG-16. We also repurpose VGG-16 and VGG-19~\cite{simonyan2014very} by global average pooling the Conv 5 features and using a 512 dimensional fully-connected layer along with a 3-way classifier. All the convolutional layers in VGG-16 and VGG-19 are initialized from VGG models trained on Imagenet~\cite{deng2009imagenet}. The state-of-the-art object recognition classifiers, Resnet-50 and Resnet-101~\cite{he2015deep} are also evaluated. We replace the final 1000-way classifier layer with a 3-way classifier. For ResNet training, we do not update the Batch Normalization (BN) layers. The BN statistics from ResNet model trained for ILSCVRC challenge~\cite{deng2009imagenet} are retained during fine-tuning for crowd-counting. The BN layers behave as a linear activation function with constant scaling and offset. We do not update the BN layers as we use a batch size of $1$ during SGD and the BN parameter update becomes noisy. 

We train each of the classifier on image patch-label pairs, with labels generated from the differential training stage (see Section 3.3 of the main paper). The classifiers are trained using SGD in a stand-alone manner similar to Section \ref{a1}.
Table \ref{t2} shows the performance of the different switch classifiers on Part A of the ShanghaiTech dataset~\cite{zhang2016single}. CNN-small shows a 10\% drop in classification accuracy over the other classifiers as it is unable to model the complex multichotomy inferred from differential training. We observe that the performance plateaus for the other classifiers despite using more powerful classifiers like ResNet. This can be attributed to complexity of manifold inferred from differential training. Hence, we choose the repurposed VGG-16 model for all our Switch-CNN experiments as it gives classification accuracy competitive with deeper models like ResNet, but with a lower computational cost. A lower computational cost is critical as it allows faster training during coupled training of the switch-classifier and CNN regressors $R_{1-3}$.


\section{Conclusion}
In this paper, we propose switching convolutional neural network that leverages intra-image crowd density variation to improve the accuracy and localization of the predicted crowd count. We utilize the inherent structural and functional differences in multiple CNN regressors capable of tackling large scale and perspective variations by enforcing a differential training regime. Extensive experiments on multiple datasets show that our model exhibits state-of-the-art performance on major datasets. Further, we show that our model learns to group crowd patches based on latent factors correlated with crowd density.

\section{Acknowledgements}
This work was supported by SERB, Department of Science and Technology (DST), Government of India (Proj No. SB/S3/EECE/0127/2015).


{\small
\bibliographystyle{ieee}
\bibliography{egbib}
}

\end{document}